\documentclass[smallextended]{svjour3}       
\smartqed  

\usepackage{amsmath}
\usepackage[colorlinks]{hyperref}
\usepackage{graphicx}
\usepackage{graphics}
\usepackage{verbatim}
\usepackage{float}
\usepackage{mathtools}
\usepackage[hyphenbreaks]{breakurl}
\usepackage{mathptmx}
\usepackage{fullpage}

\makeatletter
\g@addto@macro{\UrlBreaks}{\UrlOrds}
\makeatother

\begin{document}

\title{Construction of confidence interval for a univariate stock price signal predicted through Long Short Term Memory Network} 

\titlerunning{CI-in-AI}        
\author{ Shankhyajyoti De, Arabin Kumar Dey and Deepak Gauda
}

\institute{Shankhajyoti De \at
             Department of Mathematics,\\ 
             IIT Guwahati,\\   
             Guwahati, India\\
             \email{arabin@iitg.ac.in} \\
             \and
             Arabin Kumar Dey \at
             Department of Mathematics,\\ 
             IIT Guwahati,\\   
             Guwahati, India\\
             \email{arabin@iitg.ac.in} \\
             \and
             Deepak Gouda \at
             Department of Mathematics,\\ 
             IIT Guwahati,\\   
             Guwahati, India\\
             \email{arabin@iitg.ac.in} \\
}

\maketitle

\begin{abstract}

 In this paper, we show an innovative way to construct bootstrap confidence interval of a signal estimated
based on a univariate LSTM model. We take three different types of bootstrap methods for dependent set up. We
prescribe some useful suggestions to select the optimal block length while performing the bootstrapping of the
sample. We also propose a benchmark to compare the confidence interval measured through different bootstrap
strategies. We illustrate the experimental results through some stock price data set.

\end{abstract}

\section{Introduction}
\label{intro}
  
 One of the common drawbacks of all machine learning techniques used in Signal Processing is that none of the methods suggests any convenient way to construct the confidence intervals of the predicted signals.  Any forecasting or an estimation technique used for the prediction of a signal always leads to some level of random variations in the values of the prediction at each time point.  A statistical way to address such variation is by providing a confidence band of that predicted values.  All statistical signal processing techniques address the issue while making their prediction.  Machine learning approaches do not have any well-studied framework to address this critical problem.  One of the reasons is such construction of confidence interval requires appropriate modeling of signal noise or asymptotic properties of the estimated parameters.  In this paper, we address an innovative way to construct the confidence interval through different types of bootstrap.  Finding the best-constructed confidence interval also requires an effective measure to guess the smallest confidence band given a particular probability.  We set some criteria to address the issue, which is capable of comparing different methods.    
  
  Predicting the price of a stock is a challenging problem attempted for years.  A popular method to predict the stock is technical analysis (\cite{PatelShahThakkarKotecha:2015}, \cite{SenChaudhuri:2017}).  However plenty of different other methods based on computational intelligence accomplish accurate predictions on stock market [e.g. \cite{ZhangZhongDongWangWang:2019}, \cite{SongLeeLee:2019}, \cite{SenChaudhuri:2017}, \cite{PatelShahThakkarKotecha:2015}].  These algorithms include evolutionary computation through the genetic algorithm to support vector machine or various neural network-based approach [\cite{SainiSharma:2019}, \cite{TangDongShi:2019}], deep learning methods such as deep belief net coupled with multi-linear perceptron [\cite{LongLuCui:2019}, \cite{KhoaSakakibaraNishikawa:2006}, \cite{JiangTangChenWangHuang:2018}, \cite{SelvamuthuKumarMishra:2019}].  However, these models do not deal with the dependent stream of data.  LSTM based model is an alternative popular method \cite{HochreiterSchmidhuber:1997} used for modeling dependent data.  LSTM and its variations are useful in areas such as Natural Language Processing [\cite{ShiTangLong:2019}, \cite{ShiTangCui:2018}], Financial Time series, or some application of speech processing.  Many researchers (\cite{JiangTangChenWangHuang:2018}, \cite{ChenZhouDai:2015}, \cite{Jia:2016}) used an LSTM model to predict the stock price.  None of the literature contains any methodology/convenient way to construct an appropriate confidence band in this setup.  Also, there is no appropriate notion available to find the best confidence interval of the forecast signals in an LSTM type of model set up.

 We organize the paper as follows.  In Section 1, we provide the details of the data used in this paper. The description of the LSTM model used in this paper is available in Section 2.  We describe different types of bootstrap algorithms used for dependent set up in Section 3. In section 4, we discuss the proposed methods to construct confidence intervals in various ways and also the techniques to get the best-constructed confidence interval. The performance of the proposed architectures is available in Section 5.  We conclude the paper in Section 6. 

\section{Data Set Description}  We take two different data sets to carry out our empirical analysis.  We study S\& P 500 from 28th April 2015 to 27th April 2020, and Google stock price from 8th February 2013 to 7th February 2018.  The sample size considered in both cases is 1259. We consider the only closing price of the stock.  In an LSTM network, we use input-output values as log-return data.  However, we can recalculate the predicted closing price from log-return data. The main objective of the paper is not an innovation in predicting accurate prices; instead, the paper explores the best confidence band of the predicted stock price via an LSTM model or similar machine learning model. We divide the data set into nearly 64\% and 36\% i.e., we take the number of training data set as 800 and testing set as 459.    

\section{Usual Machine learning Algorithms used for Predictions}  

\subsection{Long Short-Term Memory (LSTM)}

  Here we introduce the construction of a special kind of cell in LSTM, which took the place of the traditional hidden unit \cite{HochreiterSchmidhuber:1997}. This cell takes in the current input $x_t$, the previous output $h_{t-1}$, and the previous cell state (memory) $c_{t-1}$. Inside the cell, there are three gates, forget gate ($f$), input gate ($i$), and output gate ($o$).  The orientation of LSTM enables the important memory of the previous time to carry forward a long way with an intervention of a forget gate at each time step, which helps to keep only the relevant information or memory.   



  The following equations govern the calculation of the gate vectors:
\begin{equation}
    f_{t} = \sigma(W_fx_t + U_fh_{t-1} + b_f)
\end{equation}
\begin{equation}
    i_{t} = \sigma(W_ix_t + U_ih_{t-1} + b_i)
\end{equation}
\begin{equation}
    o_{t} = \sigma(W_ox_t + U_oh_{t-1} + b_o).
\end{equation}

  The new values of cell state $c_t$ and output $h_t$ (along with an intermediatary variable $\tilde{c}_t$) are calculated by the following equations:
\begin{equation}
    \tilde{c}_t = \textrm{tanh}(W_cx_t + U_ch_{t-1} + b_c)
\end{equation}
\begin{equation}
    c_{t} = i_t \odot \tilde{c}_t + f_t \odot c_{t-1}
\end{equation}
\begin{equation}
    h_{t} = o_t \odot \textrm{tanh}(c_t).
\end{equation}

  Here, $\odot$ denotes Hadamard product.  

\subsection{Some Issues in LSTM model set up}  

  Window feature scaling enhances training because it enables the model to capture the \textbf{relative} variance in the data in a small neighborhood.  The feature scaling is a crucial part of training LSTM models in this context. It normalizes the entire data, feature-wise, to convert it to values between 0 and 1 using min-max scaling.  Different types of normalization are available in the literature.  However in this paper we use the following relation for normalization : $x_{norm} = \frac{x - x_{min}}{x_{max} - x_{min}}.$  Since we deal with a constant time step (lookback) while training LSTM, it makes sense to normalize the data in each lookback-length window.  However, the prediction from the model is later de-normalized.  We use the Adam optimization algorithm to train the LSTM model. The choice of optimal lookback and batch size are made by trial and error method so that the optimal number minimize RMSE of the test set. Details of choice for the parameter tuning are available in Section 5.

\section{Three different Bootstrap for dependent set up}  There are multiple methods of bootstrap available in the literature [e.g. \cite{KreissaPaparoditis:2011}, \cite{RadovanovMarcikić:2014}].  We choose the following major three methods in this paper for comparison purposes.  

\subsection{Method 1 : Non-overlapping Block Bootstrap (NBB)} We take up the following steps :

 Let $l \in N$, $l < < n$, $L = \lceil \frac{n}{l} \rceil$ and $k = n - Ll$.  Define discrete independent random variables $i_{1}, i_{2}, i_{3}, \cdots, i_{L}$ taking values in the set $I_{n, I}$, where 

\begin{enumerate}


\item $I_{n , I} = \{ 1, l + 1, 2l + 1, \cdots, (L - 1)l + 1 \}$ if non-overlapping blocks are considered.

\item Lay the blocks $(X_{i_{s}}, X_{i_{s} + 1}, \cdots, X_{i_{s} + l - 1})$, $s = 1, 2, 3, \cdots, L$ end to end in the order sampled together and discard the last $(l - k)$ observations to form a bootstrap pseudo series $X^{*}_{1}, X^{*}_{2}, \cdots, X^{*}_{n}$.
 
\end{enumerate}   

\subsection{Method 2 : Moving Block Bootstrap (MBB)} Assume $l$ is an integer between 1 to $n$ holding a property that $\lim_{l \rightarrow \infty}^{}\lim_{n \rightarrow \infty}^{} \frac{l}{n} \rightarrow 0$.  Anyway, a specific description of this method must start from considering the block length $l$ as constant.  If $B_{i} = (X_{i}, \cdots, X_{i + l - 1})$ denotes an i-th block of time series, then, the block length begins with $X_{i}$, for $1 \leq i \leq N$, where $N = n - l + 1$ is represented as the number of blocks within the bootstrap sample.  In order to form a sample from moving block bootstrap method, it is necessary to choose randomly a certain number of blocks from the set $\{ B_{1}, B_{2}, \cdots, B_{N} \}$.  Therefore $B^{*}_{1}, B^{*}_{2}, \cdots, B^{*}_{k}$ represents a random sample with repetitions from the set $\{ B_{1}, B_{2}, \cdots, B_{N} \}$ where each block contains same $l$ number of observations.     

\subsection{Method 3 : Local Block Bootstrap (LBB)}  Given the data set $X_{1}, X_{2}, \cdots, X_{n}$,  LBB algorithm creates a pseudo-series $X^{*}_{1}, X^{*}_{2}, \cdots, X^{*}_{n}$ as follows :

\begin{enumerate}

\item Select an integer block size $l$, and a real number $B \in (0, 1]$ such that $nB$ is an integer, both l and B are a function of $n$.

\item For $m = 0, 1, \cdots, (\lceil n/l\rceil - 1)$ let $X^{*}_{ml + j} = X_{I_{m} + j - 1}$ for $j = 1, 2, \cdots, l$ where $I_{1}, I_{2}, \cdots$ are independent integer valued random variables satisfying $P(I_{m} = k) = W_{n, m}(k)$. A choice $W_{n, m}(k) \sim Unif(J_{1, m}, J_{2, m})$, $J_{1, m} = \max\{ 1, ml - nB \}$, $J_{2, m} = \max\{ n - l + 1, ml + nB \}$.

\end{enumerate}
 
\section{Proposed Method:}  In this paper, we contribute in different ways.  First, we describe some implementation issues in practice to obtain the optimal bootstrap block length in this setup.  Second, we develop a method to construct a confidence interval for the stock price signal received from the LSTM network.  Third, we develop a benchmark to judge the best confidence interval in a different setup.  

\subsection{Benchmark set for best choice of the block length in Bootstrap:}  

  The choice of block length in Bootstrap is an interesting problem. Hall et al. \cite{HallHorowitzJing:1995} established that the best block size depends on three factors: the autocorrelation structure, the series length, and the reason for bootstrapping. Most of the papers discussed such choices in the context of estimation or finding sampling distribution of bootstrap mean or variances for dependent set up [\cite{NordmanLahiri:2014}, \cite{PolitisWhite:2004}, \cite{Lahiri:1999}]. The purpose of any bootstrap is to regenerate the sample, which behaves like the original sample.  Therefore it is expected that a regenerated sample should be the best representation of the original sample.  Thus the best bootstrapping algorithm mimics the data in the best possible way.  Statistically, this is possible when the variance of the generated curve shows a minimal variation for the original curve.  However, quantifying this variance is not unique, and thus many algorithms are available.  Finding the mean series from the blocks of the original and bootstrapped sample and creating a distance function for the same could be a way to quantify this variation.  Demirel and Willemain (\cite{DemirelWillemain:2002}) uses a similar concept in their paper to find the best bootstrap length. His approach uses Higher-order crossings (HOC) to form the benchmark to find out the bootstrap length.  This approach does not have a proper theoretical basis for non-stationary data and has many shortcomings. 

  In our paper, we proposed a novel penalty function-based approach to select the block length. Instead of a HOC based statistic, we proposed using the following the empirical distance function based on the mean series of the original sample and bootstrapped sample :
$\frac{1}{M} \sum_{i = 1}^{M} \frac{l}{n}\sum_{T = 1}^{b} (\bar{X}^{*}_{T, i} - \bar{X}_{T, i})^{2}$, where $M$ : the number of bootstrap sample drawn, $n$ : the series length, $b$ : the number of blocks, $l$ : the block length and $\bar{X}^{*}_{T, i}$, $\bar{X}_{T, i}$ are mean series of bootstrapped and original sample respectively.  Based on our intuition, we can realize that the best block length should be the length, which minimizes the above empirical function.  

\begin{figure}[ht!]
\begin{center}
    \includegraphics[width = 0.9\textwidth]{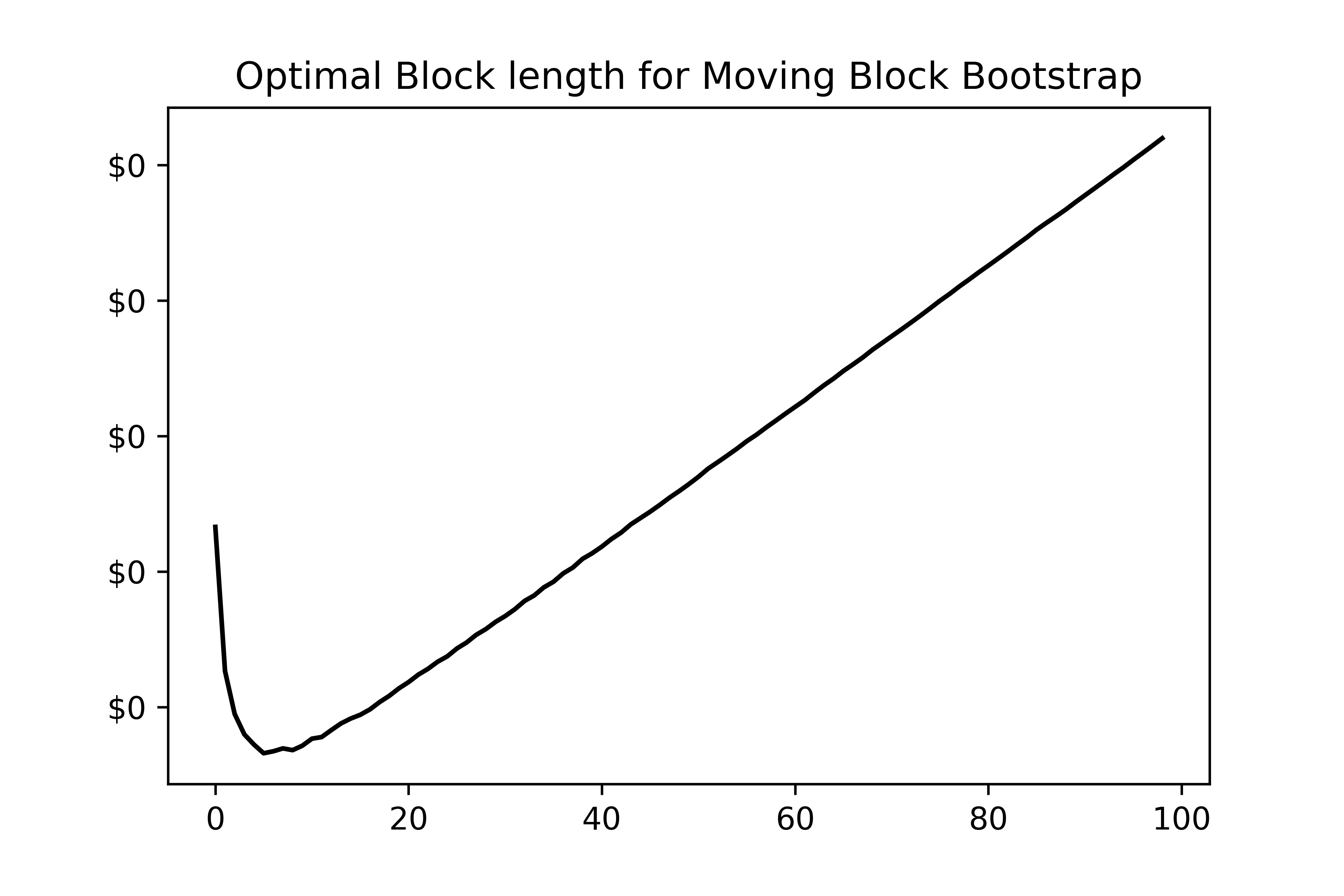}
\caption{Optimal Block length selector in Moving Block Bootstrap for Google stock price \label{Figure8}}
\end{center}
\end{figure}

\begin{figure}[ht!]
\begin{center}
    \includegraphics[width = 0.9\textwidth]{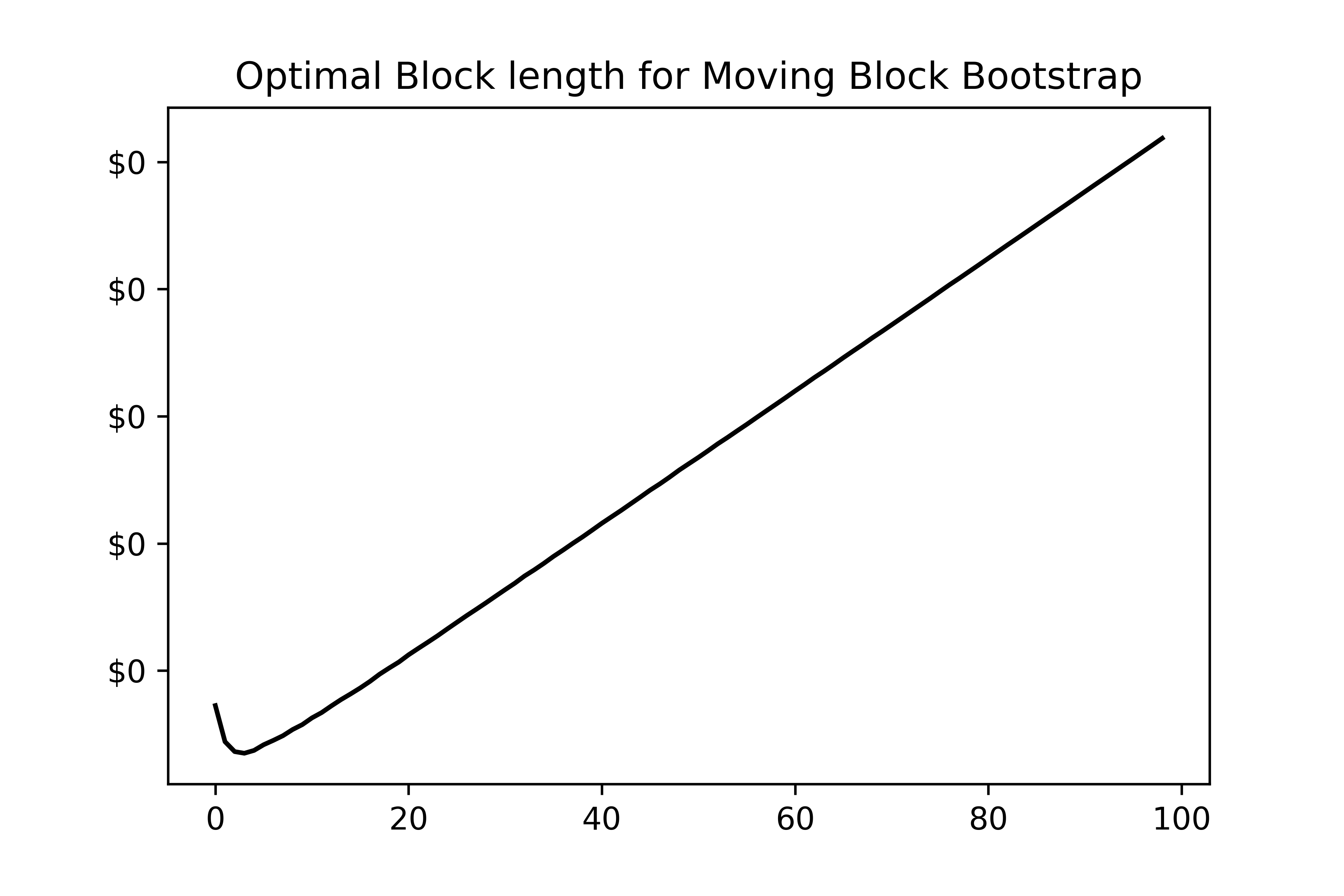}
\caption{Optimal Block length selector in Moving Block Bootstrap for S\&P 500 stock price \label{Figure9}}
\end{center}
\end{figure}

 However we observe numerically that the above empirical function decreases with lots of oscillations with respect to block length. As suggested by Hall et al. \cite{HallHorowitzJing:1995}, this oscillation depends on the autocorrelation structure, the series length which makes selection of block length very difficult.  If we plan to use a penalty function in addition to this empirical function, convexity of the function depends again with suitable choice of the series length. A separate attention/research is required in this direction. However we propose to use a different trick in this context.  We propose to take block length which minimizes the following function :
$\frac{1}{M} \sum_{i = 1}^{M} \frac{l}{n}\sum_{T = 1}^{b} (\bar{X}^{*}_{T, i} - \bar{X}_{T, i})^{2} + \frac{\log(n)}{n^t} l$, where $t = 2$.
 And $\bar{X}^{*}_{T, i}$, $\bar{X}_{T, i}$ are mean series of original data and bootstrapped data when we take original data as log-returns of the prices. This small modification makes the function an excellent convex function for block length.  We may need to adjust the value of $t$ for exceptional data, as this determines the convergence rate of the above the selector of the block length.  However, from our experimental analysis and empirical evidence, we observe it is reasonable to take its value as 2.  Fig.\ref{Figure8} and Fig.\ref{Figure9} reveal the pictorial overview of the convexity of the block length selector in Moving Block Bootstrap when the value of $t$ is 2. The figures for other procedures are similar; therefore, we omit them from the paper.  They can be made available on request to authors.
 
  The above procedure does not create any problem to fulfill our objective of constructing a confidence interval.  We can graphically observe that conversion of the original data into a log-return series, and a reverse transformation of the bootstrapped log-return data is capable of mimicking the original data equally well. The above procedure is simple to implement and can be applied even when Higher-order crossings (HOC) fails to capture the block size accurately for a direct, highly volatile, and non-stationary type of data.  However, our proposed approach needs more systematic analysis.  We pursue the details in some other work. 

\subsection{Methodology for Construction of Confidence Interval: } Our proposed method to construct the confidence interval in LSTM set up is as follows :
\begin{enumerate}
\item Create bootstrap samples, which mimic the original sample of the input.
\item For each bootstrap sample, fit an LSTM model separately and predict the test set.     
\item At each time step, find out 95\% non-parametric confidence interval to construct the confidence interval of the test set.
\end{enumerate}

 In this paper, we propose to construct a slightly different bootstrap sample for closing prices instead of directly extracting bootstrapped samples.  We observe that our algorithm depends on the choice of optimal block length that requires transforming the data into the log-return series.  Therefore we first transform the data into log-returns and choose optimal block length based on log-return series.  Different bootstrap techniques are used on the log-return data using the calculated optimal block length. A reverse transformation helps to create the bootstrap sample for closing prices.  We observe graphically that such bootstrap data mimics the original data also very well. 

\subsection{Benchmark:}  Setting a benchmark is crucial in comparing the constructed confidence intervals.  We use the sum of confidence width at each time instance as a naive approximation of confidence band, which we consider as comparing factor.  We calculate this comparing factor for test set price signal for each bootstrap/or other procedures and use this quantity as a benchmark for comparing different confidence intervals obtained by various methods.   The best-constructed confidence interval would be the one providing a minimum sum of confidence width.  

\section{Performance of the Architectures and Numerical Results}

\subsection{Parameter Tuning}  While applying the LSTM, we take batch size as 15 and look back number as 5.  This choice is made by trial and error, minimizing the RMSE.   We take the smoothing window as 200, whereas we take the dropout rate at 0.2.  Again optimal no of epochs is kept as 19 since RMSE does not have significantly lower value after 19. Finally, training and testing set RMSE is found at 0.1296 and 25.2169, respectively.  We introduce kernel regularizer, which is essentially regularization on the weights of neurons.  The optimal choice of block length depends on the dataset we use for empirical analysis. For S\&P 500 data, we get block length as 3 for LBB and 4 for NBB and MBB.  The optimal block length for Google stock price data is 6 for all bootstrap algorithms.  We make all empirical analyses based on 1000 generated bootstrapped samples.

\subsection{Graphical Sense of Bootstrap data :}  
\begin{figure}[ht!]
\begin{center}
    \includegraphics[width = 0.9\textwidth]{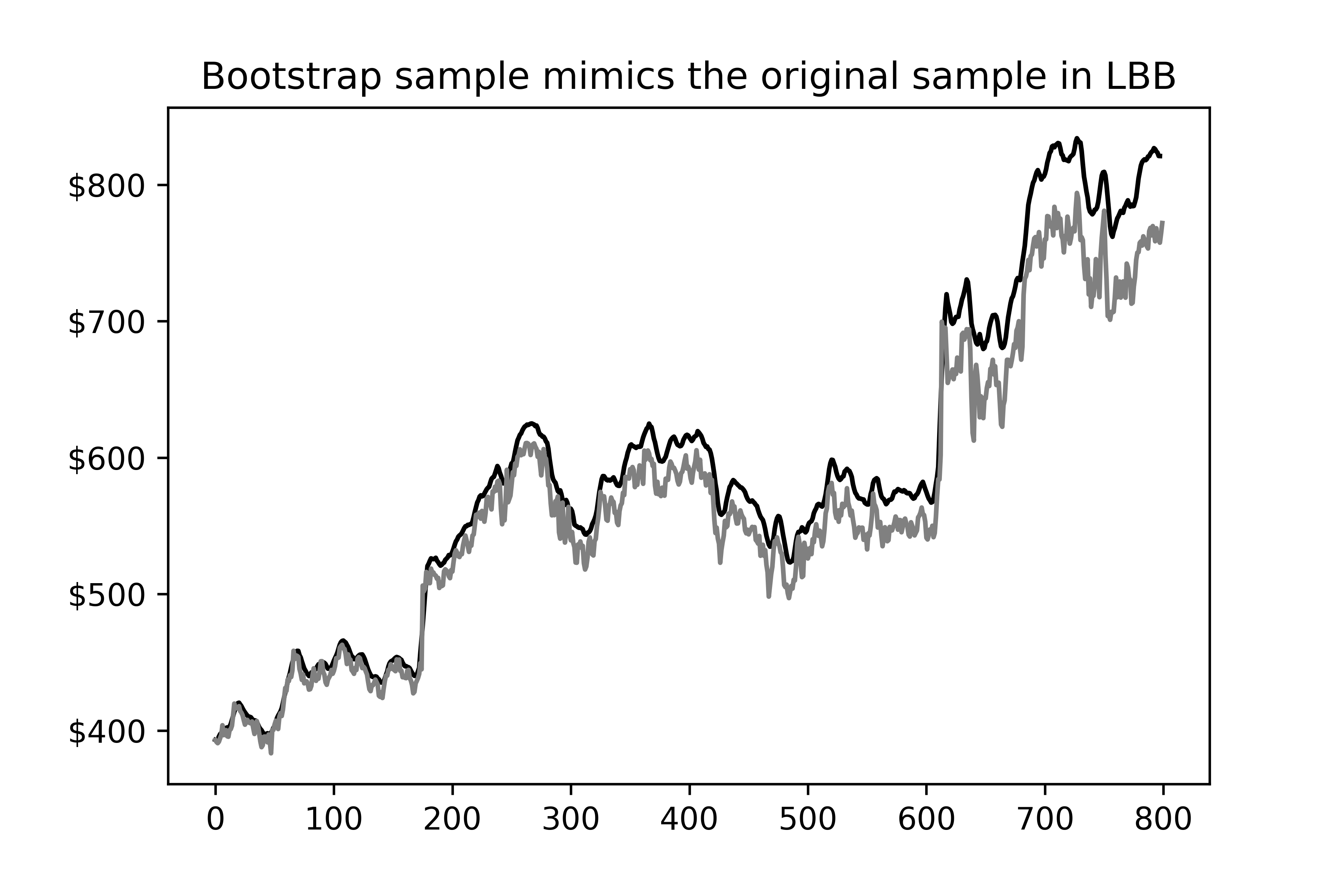}
\caption{\label{Figure.1} Figures showing how LBB method mimics original data using Google stock price}
\end{center}
\end{figure}

\begin{figure}[ht!]
\begin{center}
    \includegraphics[width = 0.9\textwidth]{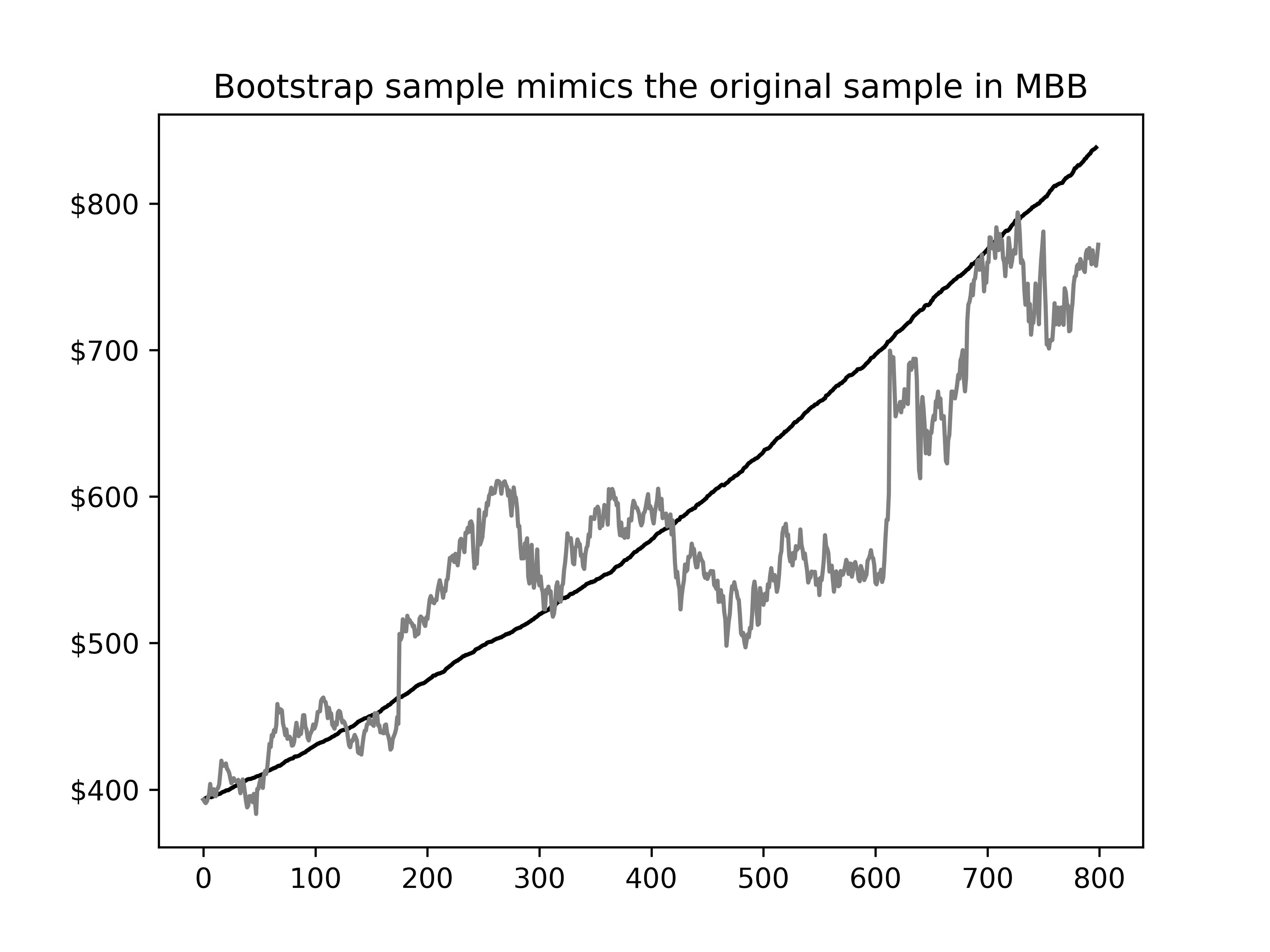}
\caption{Figures showing how MBB method mimics original data using Google ptock Price\label{Figure.2}}
\end{center}
\end{figure}

\begin{figure}[ht!]
\begin{center}
    \includegraphics[width = 0.9\textwidth]{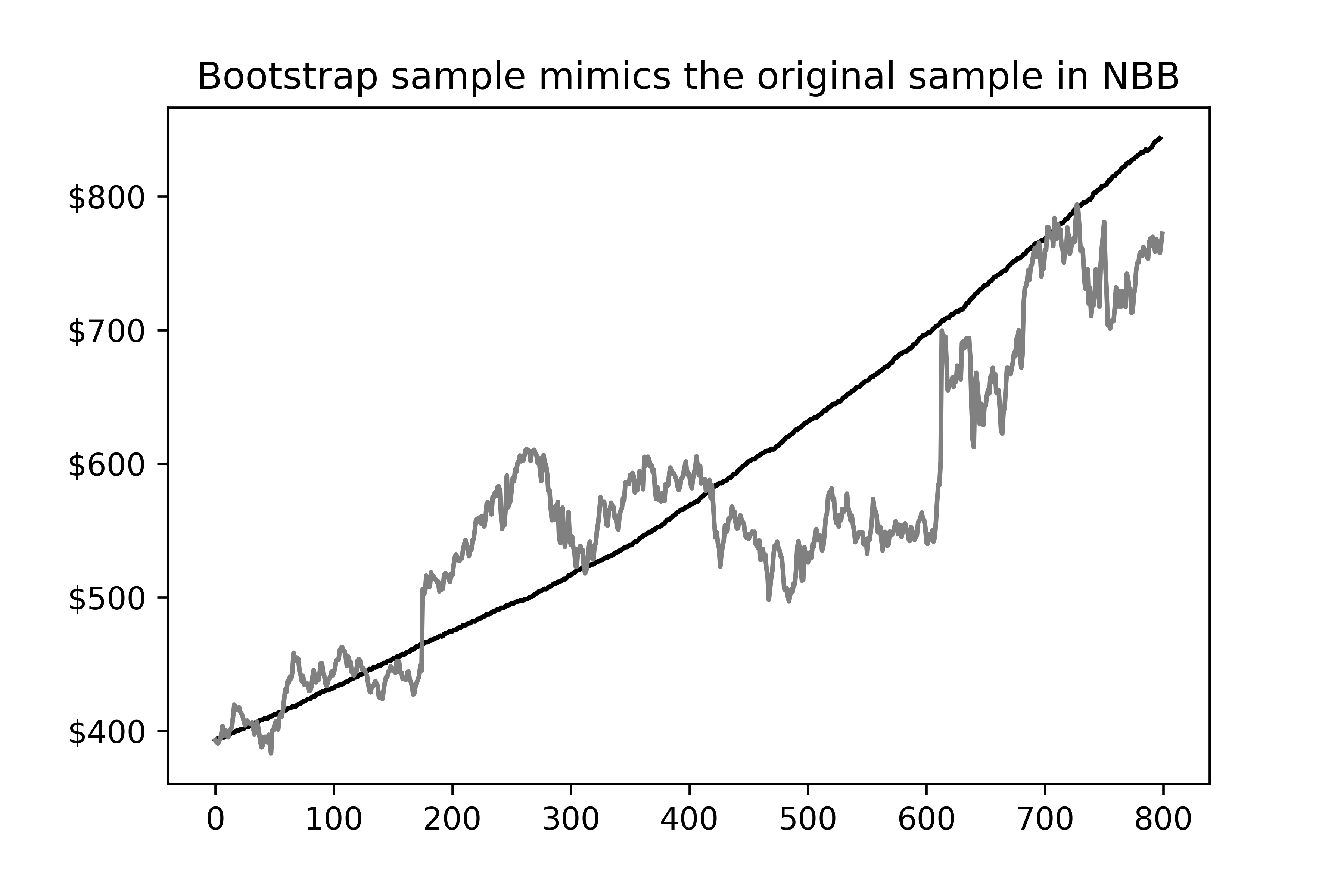}
\caption{Figures showing how NBB method mimics original data using Google stock price \label{Figure.3}}
\end{center}
\end{figure}

It is also important to figure out how the optimal choice of bootstrap block length enables the resampling technique to mimic the original data set.  We take a graphical sense of the fact by plotting original data and bootstrap resampled data obtained through different procedures. Fig.\ref{Figure.1}, Fig.\ref{Figure.2}, Fig.\ref{Figure.3} show that local block bootstrap best captures the patterns of the original data.

\subsection{Results and Comparisons on the Data set}  
\begin{figure}[ht!]
\begin{center}
    \includegraphics[width = 1.0\textwidth]{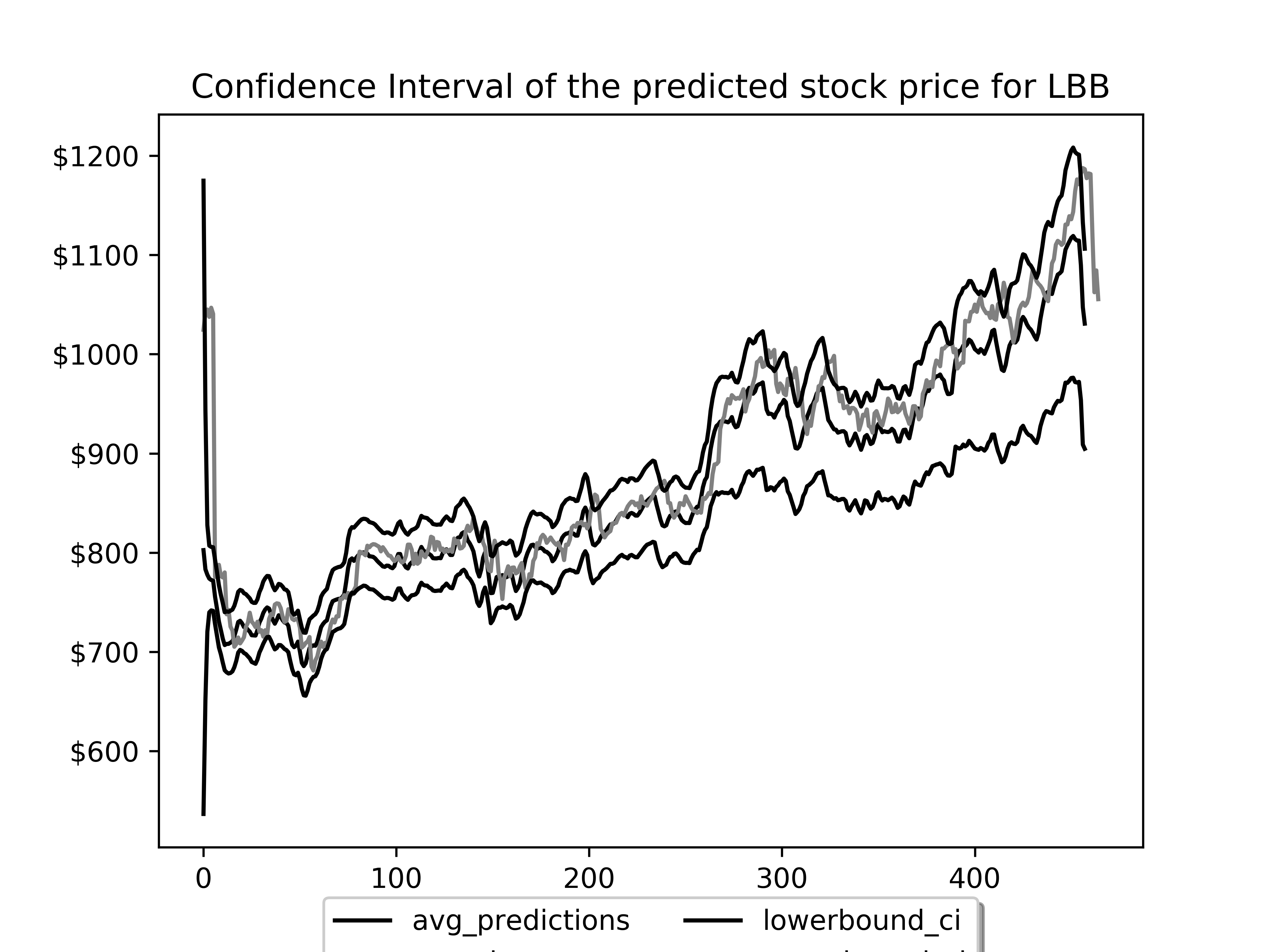}
\caption{Confidence Interval obtained through LBB on LSTM model using Google stock price \label{Fig.1}}
\end{center}
\end{figure}

\begin{figure}[ht!]
\begin{center}
    \includegraphics[width = 1.0\textwidth]{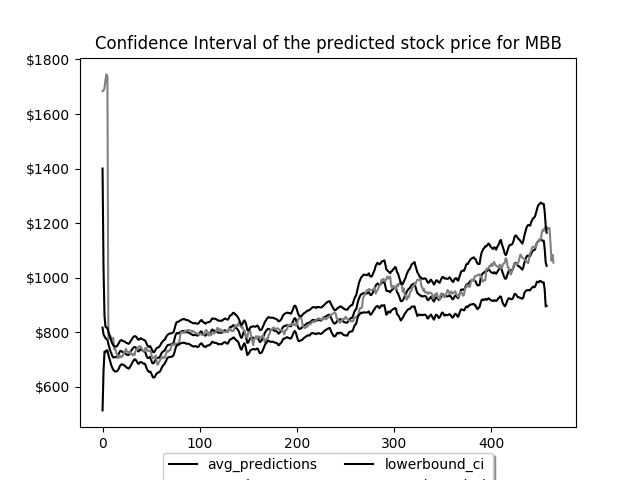}
\caption{Confidence Interval obtained through MBB on LSTM model using Google stock price\label{Fig.2}}
\end{center}
\end{figure}

\begin{figure}[ht!]
\begin{center}
    \includegraphics[width = 1.0\textwidth]{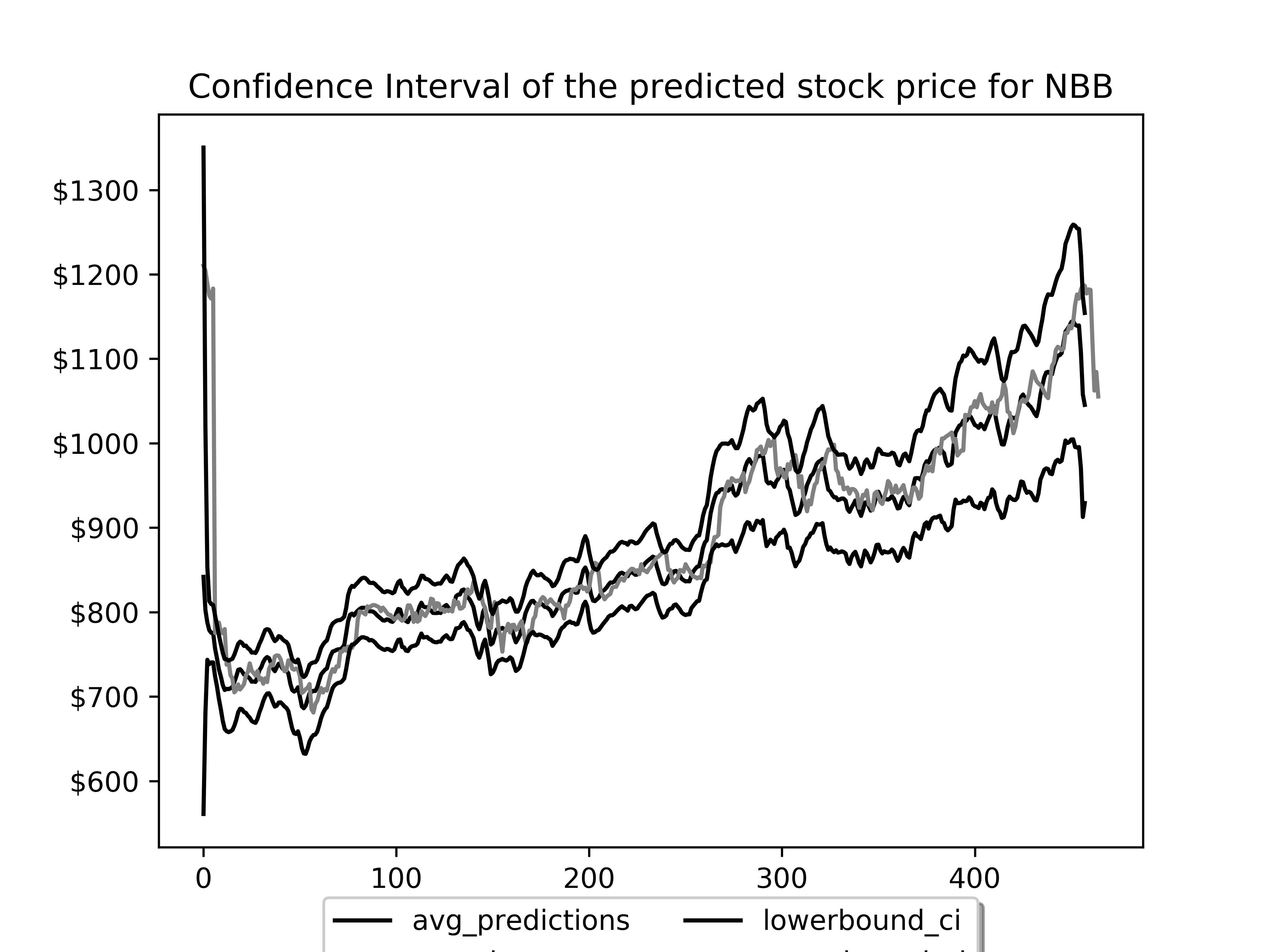}
\caption{Confidence Interval obtained through NBB on LSTM model using Google stock price \label{Fig.3}}
\end{center}
\end{figure}

\begin{figure}[ht!]
\begin{center}
    \includegraphics[width = 0.9\textwidth]{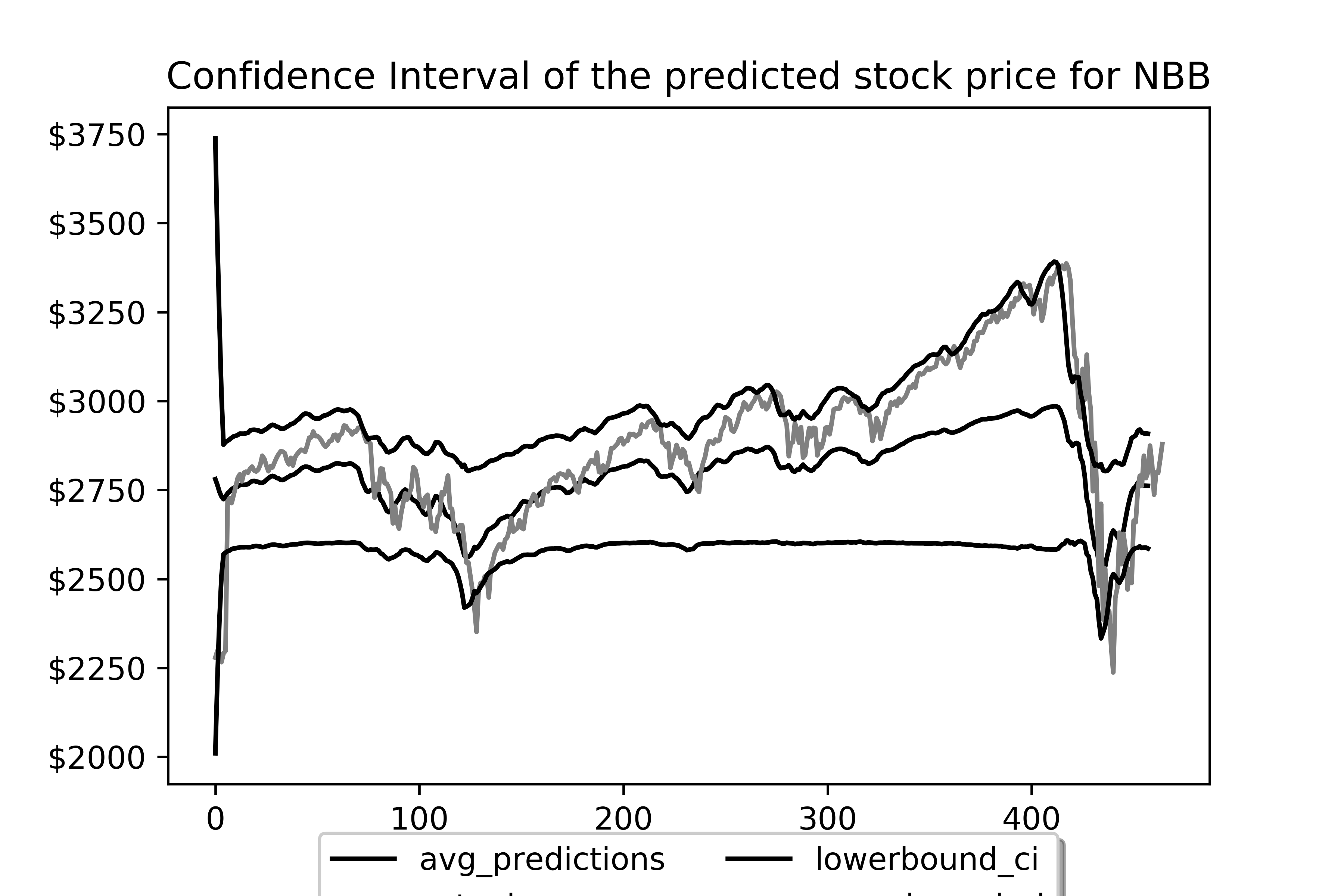}
\caption{Confidence Interval obtained through NBB on LSTM model using S\&P 500 stock price \label{Fig.5}}
\end{center}
\end{figure}

\begin{figure}[ht!]
\begin{center}
    \includegraphics[width = 0.9\textwidth]{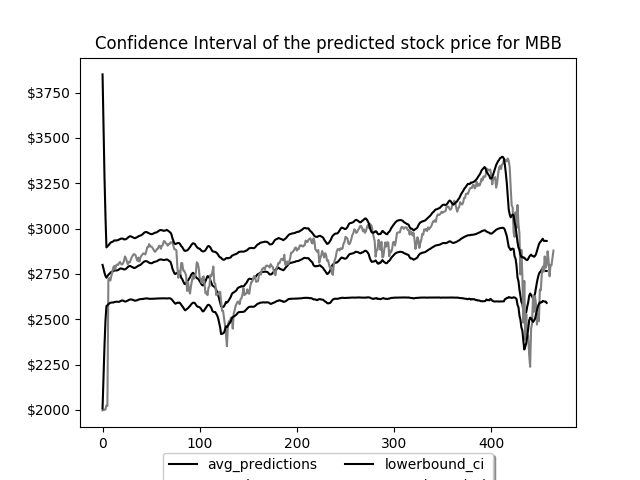}
\caption{Confidence Interval obtained through MBB on LSTM model using S\&P 500 stock price \label{Fig.6}}
\end{center}
\end{figure}

\begin{figure}[ht!]
\begin{center}
    \includegraphics[width = 0.9\textwidth]{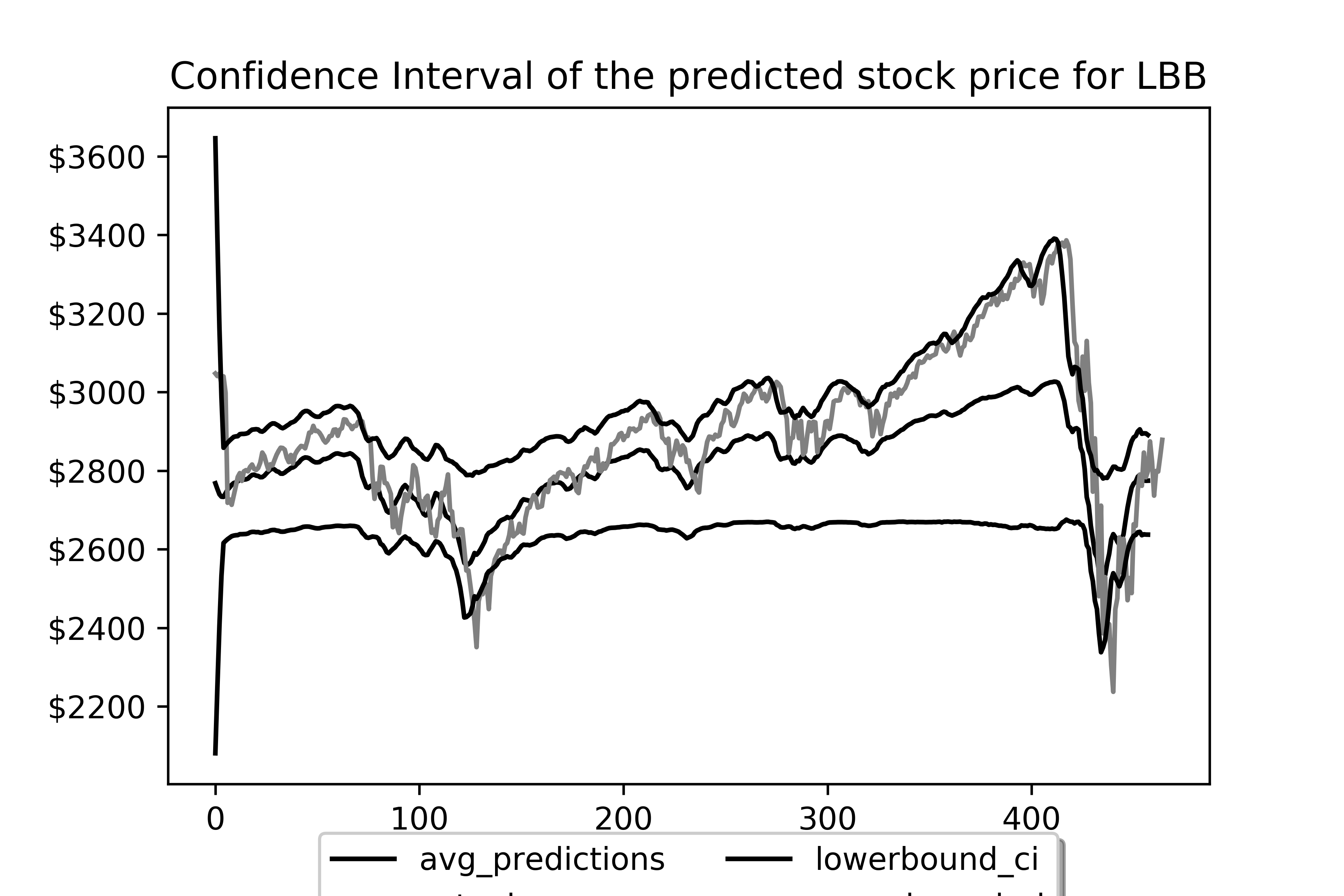}
\caption{Confidence Interval obtained through LBB on LSTM model using S\&P 500 stock price \label{Fig.7}}
\end{center}
\end{figure}

Fig.\ref{Fig.1}, Fig.\ref{Fig.2}, Fig.\ref{Fig.3} provide the picture of the constructed confidence interval of Google stock price on the test set by Local block bootstrap, Non-overlapping block bootstrap, Moving block bootstrap, respectively.  For S\& P 500, we see the prediction bands in Fig.\ref{Fig.5}, Fig.\ref{Fig.6}, Fig.\ref{Fig.7} respectively.  The Prediction in the case of S\& P 500, deals with a lesser number of hidden parameters within LSTM nodes.  We observe that the predicted curve better matches with actual observations if we increase the number of hidden nodes in each LSTM cell. As S\& P 500 includes some recent stock price data, it reflects the behavior in the presence of the current coronavirus pandemic.  The last part of the Figure shows a sharp depletion of the stock price, which is the pandemic effect.  We can see how the confidence interval constructed captures original data in the presence of the coronavirus shock wave.  We also compute the proposed comparing factor for both the data using the three of these bootstrap procedures.  For Google stock price, we obtain the values as 47484.094, 53325.1136, and 53767.2914, respectively, whereas for S\& P 500, we get the numbers as 163697.7224, 184850.4583, 189109.1758 respectively. As the minimum comparing factor should provide the best procedure for the setup, Local block bootstrap appears to be the best bootstrap procedure to construct a confidence interval in univariate LSTM set up which comes out as the best in terms of comparing factor in S\&P 500 as well as Google stock price data set.  It makes sense to get Local block bootstrap as the best procedure in all cases, as we observe from Fig.\ref{Figure.1} that it mimics the original data the best as compared to other algorithms.

\section{Conclusion}  We verify graphically that the proposed choice of Bootstrap length is capable of mimicking the data set very well.  Local block bootstrap has the best ability to mimic the original data set.  We tune parameters in such a way that the LSTM model becomes capable of predicting the stock price efficiently.  Three bootstrap procedures work quite well to determine a long term confidence band.  We found local block and non-overlapping block bootstrap performs almost equally well and better than the moving block bootstrap method.  We can also extend the idea in the Spatio-temporal setup or in a multivariate time series model.  The work is in progress.      

\bibliographystyle{spmpsci}
\bibliography{boot}

\begin{thebibliography}{10}
\providecommand{\url}[1]{{#1}}
\providecommand{\urlprefix}{URL }
\expandafter\ifx\csname urlstyle\endcsname\relax
  \providecommand{\doi}[1]{DOI~\discretionary{}{}{}#1}\else
  \providecommand{\doi}{DOI~\discretionary{}{}{}\begingroup
  \urlstyle{rm}\Url}\fi

\bibitem{ChenZhouDai:2015}
Chen, K., Zhou, Y., Dai, F.: A lstm-based method for stock returns prediction:
  A case study of china stock market.
\newblock In: 2015 IEEE international conference on big data (big data), pp.
  2823--2824. IEEE (2015)

\bibitem{DemirelWillemain:2002}
Demirel, O.F., Willemain, T.R.: Generation of simulation input scenarios using
  bootstrap methods.
\newblock Journal of the Operational Research Society \textbf{53}(1), 69--78
  (2002)

\bibitem{HallHorowitzJing:1995}
Hall, P., Horowitz, J.L., Jing, B.Y.: On blocking rules for the bootstrap with
  dependent data.
\newblock Biometrika \textbf{82}(3), 561--574 (1995)

\bibitem{HochreiterSchmidhuber:1997}
Hochreiter, S., Schmidhuber, J.: Long short-term memory.
\newblock Neural computation \textbf{9}(8), 1735--1780 (1997)

\bibitem{Jia:2016}
Jia, H.: Investigation into the effectiveness of long short term memory
  networks for stock price prediction.
\newblock arXiv preprint arXiv:1603.07893  (2016)

\bibitem{JiangTangChenWangHuang:2018}
Jiang, Q., Tang, C., Chen, C., Wang, X., Huang, Q.: Stock price forecast based
  on lstm neural network.
\newblock In: International Conference on Management Science and Engineering
  Management, pp. 393--408. Springer (2018)

\bibitem{KhoaSakakibaraNishikawa:2006}
Khoa, N.L.D., Sakakibara, K., Nishikawa, I.: Stock price forecasting using back
  propagation neural networks with time and profit based adjusted weight
  factors.
\newblock In: 2006 SICE-ICASE International Joint Conference, pp. 5484--5488.
  IEEE (2006)

\bibitem{KreissaPaparoditis:2011}
Kreiss, J.P., Paparoditis, E.: Bootstrap methods for dependent data: A review.
\newblock Journal of the Korean Statistical Society \textbf{40}(4), 357--378
  (2011)

\bibitem{Lahiri:1999}
Lahiri, S.N.: Theoretical comparisons of block bootstrap methods.
\newblock Annals of Statistics pp. 386--404 (1999)

\bibitem{LongLuCui:2019}
Long, W., Lu, Z., Cui, L.: Deep learning-based feature engineering for stock
  price movement prediction.
\newblock Knowledge-Based Systems \textbf{164}, 163--173 (2019)

\bibitem{NordmanLahiri:2014}
Nordman, D.J., Lahiri, S.N., et~al.: Convergence rates of empirical block
  length selectors for block bootstrap.
\newblock Bernoulli \textbf{20}(2), 958--978 (2014)

\bibitem{PatelShahThakkarKotecha:2015}
Patel, J., Shah, S., Thakkar, P., Kotecha, K.: Predicting stock and stock price
  index movement using trend deterministic data preparation and machine
  learning techniques.
\newblock Expert systems with applications \textbf{42}(1), 259--268 (2015)

\bibitem{PolitisWhite:2004}
Politis, D.N., White, H.: Automatic block-length selection for the dependent
  bootstrap.
\newblock Econometric Reviews \textbf{23}(1), 53--70 (2004)

\bibitem{RadovanovMarcikić:2014}
Radovanov, B., Marciki{\'c}, A.: A comparison of four different block bootstrap
  methods.
\newblock Croatian Operational Research Review \textbf{5}(2), 189--202 (2014)

\bibitem{SainiSharma:2019}
Saini, A., Sharma, A.: Predicting the unpredictable: An application of machine
  learning algorithms in indian stock market.
\newblock Annals of Data Science pp. 1--9 (2019)

\bibitem{SelvamuthuKumarMishra:2019}
Selvamuthu, D., Kumar, V., Mishra, A.: Indian stock market prediction using
  artificial neural networks on tick data.
\newblock Financial Innovation \textbf{5}(1), 16 (2019)

\bibitem{SenChaudhuri:2017}
Sen, J., Chaudhuri, T.: A robust predictive model for stock price forecasting.
\newblock In: Proceedings of the 5th International Conference on Business
  Analytics and Intelligence (ICBAI 2017), Indian Institute of Management,
  Bangalore, INDIA (2017)

\bibitem{ShiTangCui:2018}
Shi, Y., Tang, Y.r., Cui, L.x., Wen, L.: A test mining based study of investor
  sentiment and its influence on stock returns.
\newblock Economic Computation \& Economic Cybernetics Studies \& Research
  \textbf{52}(1) (2018)

\bibitem{ShiTangLong:2019}
Shi, Y., Tang, Y.r., Long, W.: Sentiment contagion analysis of interacting
  investors: Evidence from china’s stock forum.
\newblock Physica A: Statistical Mechanics and its Applications \textbf{523},
  246--259 (2019)

\bibitem{SongLeeLee:2019}
Song, Y., Lee, J.W., Lee, J.: A study on novel filtering and relationship
  between input-features and target-vectors in a deep learning model for stock
  price prediction.
\newblock Applied Intelligence \textbf{49}(3), 897--911 (2019)

\bibitem{TangDongShi:2019}
Tang, H., Dong, P., Shi, Y.: A new approach of integrating piecewise linear
  representation and weighted support vector machine for forecasting stock
  turning points.
\newblock Applied Soft Computing \textbf{78}, 685--696 (2019)

\bibitem{ZhangZhongDongWangWang:2019}
Zhang, K., Zhong, G., Dong, J., Wang, S., Wang, Y.: Stock market prediction
  based on generative adversarial network.
\newblock Procedia computer science \textbf{147}, 400--406 (2019)

\end{thebibliography}


\end{document}